\documentclass[11pt,a4paper]{article}
\usepackage[hyperref]{ranlp2025}
\usepackage{times}
\usepackage{latexsym}
\usepackage{float}
\usepackage{graphicx}
\usepackage{lscape}
\usepackage{booktabs}
\usepackage{multirow}
\usepackage{longtable}
\usepackage{geometry}
\usepackage{pdflscape}
\usepackage{xcolor}
\usepackage{multirow}
\usepackage{subcaption}
\usepackage{caption} 
\usepackage{xspace}
\usepackage{placeins}  % to Force the Table to Appear Higher

\usepackage{microtype}
\usepackage{booktabs, tabularx}
\usepackage{comment}

\usepackage{etoolbox}  % Add this line to load etoolbox

\makeatletter
\patchcmd{\@bibitem}{\item}{\vspace{0pt}\item}{}{}
\patchcmd{\@lbibitem}{\item}{\vspace{0pt}\item}{}{}
\makeatother

\newcommand{\PAIlogo}{\raisebox{3.4pt}{\includegraphics[scale=0.10]{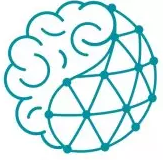}}}
\newcommand{\NICElogo}{\raisebox{3.4pt}{\includegraphics[scale=0.010]{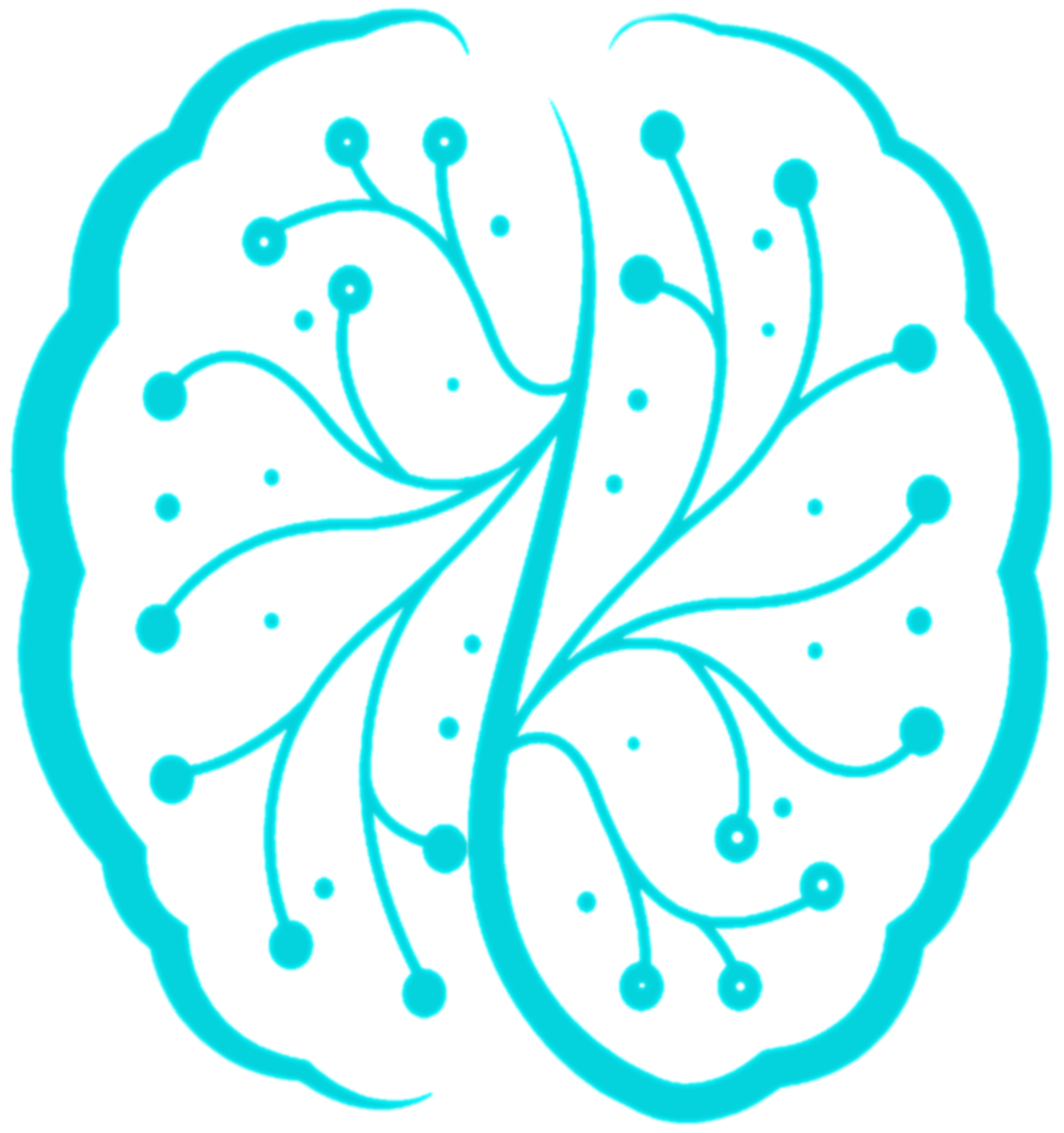}}}
\newcommand{\gemmatwob}{\texttt{Gemma-2-2B}\xspace}
\newcommand{\gemmafourb}{\texttt{Gemma-3-4B}\xspace}
\newcommand{\gemmanineb}{\texttt{Gemma-2-9B}\xspace}

\title{Cyberbullying Detection \textit{via} Aggression-Enhanced Prompting}
\author{
Aisha Saeid\NICElogo, 
Anu Sabu\NICElogo, 
Girish A. Koushik\NICElogo\\
\textbf{Ferrante Neri}\NICElogo \&
\textbf{Diptesh Kanojia}\PAIlogo \\
\NICElogo NICE Research Group \& \PAIlogo Institute for People-Centred AI,\\
School of Computer Science \& Electronic Engineering, \\
University of Surrey, UK \\
\texttt{\{a.saeid,as05318,g.koushik,f.neri,d.kanojia\}@surrey.ac.uk}
}
 
\aclfinalcopy
\date{}

\begin{document}

\maketitle

\begin{abstract}
Detecting cyberbullying on social media remains a critical challenge due to its subtle and varied expressions. This study investigates whether integrating aggression detection as an auxiliary task within a unified training framework can enhance the generalisation and performance of large language models (LLMs) in cyberbullying detection. Experiments are conducted on five aggression datasets and one cyberbullying dataset using instruction-tuned LLMs. We evaluated multiple strategies: zero-shot, few-shot, independent LoRA fine-tuning, and multi-task learning (MTL). Given the inconsistent results of MTL, we propose an enriched prompt pipeline approach in which aggression predictions are embedded into cyberbullying detection prompts to provide contextual augmentation. Preliminary results show that the enriched prompt pipeline consistently outperforms standard LoRA fine-tuning, indicating that aggression-informed context significantly boosts cyberbullying detection. This study highlights the potential of auxiliary tasks, such as aggression detection, to improve the generalisation of LLMs for safety-critical applications on social networks.

\end{abstract}

\section{Introduction}

Digital communication platforms and social media networks have fundamentally transformed human interaction, enabling expression across geographic and cultural divides. However, these platforms have also become hotspots for harmful behaviours such as cyberbullying and online harassment~\cite{hinduja2010bullying,kowalski2014bullying}. Cyberbullying, in particular, is a pressing issue due to its profound psychological and emotional impact on victims~\cite{wang2019affective}. Unlike traditional bullying, which could be confined to specific times and locations, cyberbullying victims can be targeted unchecked, as observed across multiple digital platforms, significantly impacting the young adult populace~\cite{slonje2013nature}.

Detecting cyberbullying is challenging due to its linguistic complexity, ranging from overt aggression to subtle, passive-aggressive expressions that often elude conventional moderation systems with implicit targeted meaning~\cite{xu2012bullying}. Traditional keyword-based and rule-driven detection methods struggle to adapt to evolving slang and hidden messaging patterns in social media~\cite{waseem2017understanding}. Recent advances in Large Language Models (LLMs), such as the Gemma series~\cite{gemma2024open}, have contributed to improvements in natural language processing (NLP),  particularly in understanding context, sentiment, and linguistic subtleties. However, when applied to cyberbullying detection, these models may face challenges due to domain-specific vocabulary, implicit language use, and the scarcity of annotated data tailored to harmful online content~\cite{yi2025detecting}.

This study investigates whether integrating aggression-labelled datasets with cyberbullying datasets within a unified training framework can enhance the generalisation and detection capabilities of large language models (LLMs). Moreover, we investigate architectural modifications, including targeted Transformer-heads adaptations~\cite{houlsby2019parameter}, designed to focus model attention on aggression signals, thus improving the detection of both direct and indirect cyberbullying. A core challenge addressed here is the detection of nuanced cyberbullying, especially in low-resource settings with limited annotated data. Although LLMs are versatile, their performance decreases without task-specific supervision, particularly for subtle, context-dependent aggression~\cite{chang2024survey}. Cyberbullying-only datasets often lack coverage of the entire aggression spectrum, resulting in models that miss harmful content or generate excessive false positives, undermining moderation trust~\cite{salminen2020online}. Furthermore, conventional attention mechanisms struggle to prioritise aggression signals amid noisy social media text~\cite{khan2022aggression}. 

Guided by the question of \textbf{how joint training in aggression and cyberbullying data can enhance LLM effectiveness}, our work \textit{examines how Transformer-head modifications can optimise for cyberbullying prediction using aggression-related features}. We evaluated model performance in zero-shot, few-shot, and fine-tuning scenarios, and introduce a novel prompt enrichment pipeline that leverages aggression predictions to augment cyberbullying detection.

The remainder of this article is structured as follows. Section~\ref{sec:Related} reviews related work on cyberbullying detection, aggression modelling, and the adaptation of LLMs. Section~\ref{sec:Method} outlines the methodology, including datasets, experimental design, model architectures, and evaluation metrics. Section~\ref{sec:Results} presents comparative evaluations across zero-shot, few-shot, independent LoRA fine-tuning, and multi-task learning baselines. The discussion addresses the effectiveness of the model, ethical considerations, limitations, and directions for future research. Finally, Section~\ref{sec:conc} concludes the paper by summarising the key contributions and their implications for advancing research in cyberbullying detection, and outlines potential directions for future work to further enhance the effectiveness and generalisability of LLMs in this domain.

\section{Related Work}
\label{sec:Related}

Early cyberbullying detection systems relied on rule-based and lexicon-based methods, which struggled with ambiguous language and evolving slang, often producing false positives and missing actual abuse~\cite{maity2023explain, 10.5120/ijca2025925403}. Machine learning (ML) models such as Support Vector Machines (SVMs) improved detection by incorporating sentiment and linguistic features, but still required extensive feature engineering and failed to generalize well across platforms~\cite{yi2022cyberbullying}. The advent of deep learning and LLMs like BERT, RoBERTa, and Gemma advanced the field by enabling better contextual understanding. Fine-tuned transformer models now demonstrate notable gains in precision and recall for cyberbullying detection using task-specific data~\cite{gutierrez2024improving, philipo2025assessing, gemma2024open}.
\paragraph{Role of Aggression in Cyberbullying} Recent studies show that aggression signals improve both the accuracy and interpretability of cyberbullying detection, helping models better identify hostile intent compared to general toxicity models.~\cite{han2024innovative}. Researchers have emphasised that aggression manifests across a spectrum from overt insults to subtle forms such as sarcasm and passive aggression, which often remain undetected in models trained solely on cyberbullying datasets~\cite{salminen2020classifier}. Recent empirical work by~\cite{zhao2019intervene} demonstrated that aggression-labelled datasets significantly improve models' ability to disambiguate between harmful and non-harmful content in noisy social media environments. 

\paragraph{Zero-shot and Few-shot Learning} Parallel to the development of richer datasets, zero-shot and few-shot learning paradigms have emerged as compelling solutions to data scarcity challenges, enabling models to generalise to unseen tasks. Minimal supervision-based techniques have shown notable success in sentiment analysis and hate speech detection, where labelled data are often sparse or unevenly distributed~\cite{brown2020language, schick2021exploiting}. Pre-trained LLMs such as Gemma are inherently suited to zero-shot transfer, yet their performance in cyberbullying detection without extensive fine-tuning remains inconsistent~\cite{yi2025detecting}. The effectiveness of these models improves when few-shot examples incorporate domain-specific cues like aggression indicators, as they provide contextual anchors that guide model predictions toward the correct semantic space~\cite{vidgen2019challenges}. However, most existing research does not systematically explore how multi-level aggression cues, when embedded in zero- and few-shot learning settings, can enhance cyberbullying detection.

% Recent work in model adaptation techniques, particularly Transformer-head modifications, has shown promise for improving task-specific performance in certain Large Language Models (LLMs). However, such adaptations do not universally enhance performance across all LLMs; some models exhibit significant gains while others show limited or inconsistent improvements, depending on model architecture and task complexity~\cite{houlsby2019parameter, pfeiffer2020adapterhub}. Adapter modules and projected attention layers offer more parameter-efficient alternatives to full fine-tuning, enabling models to focus on task-relevant features with minimal updates. In abusive language detection, Transformer-head adaptations have been employed to better capture subtle aggression cues~\cite{sun2022paradigm}, but their application in multi-domain cyberbullying detection remains underexplored. \\
Studies in~\cite{zampieri-etal-2020-semeval, ranasinghe2023text} demonstrate that aggression detection models can identify overt and covert hostility with reasonable accuracy. Furthermore, prompt engineering techniques have recently emerged as powerful strategies for injecting auxiliary knowledge into LLM inputs, enhancing task-specific performance without extensive re-training~\cite{zhou2022human}. However, limited research has explored using aggression detection outputs to enrich prompts specifically for improving cyberbullying detection, a gap that this study aims to address. Building upon these observations, this study proposes an integrated framework that leverages aggression-informed prompt enrichment and parameter-efficient adaptations to enhance cyberbullying detection performance, which is detailed in the following sections.

\section{Methodology}\label{sec:Method}
This section outlines the methodology used to evaluate the effectiveness of integrating aggression detection into the classification pipeline for enhancing cyberbullying detection using instruction-tuned LLMs. Our approach leverages standard classification output from the Low-rank adaptation (LoRA) adapter in a hierarchical manner for enriching cyberbullying detection. 

Our analysis focuses on three fundamental questions: (i) how well LLMs generalise to cyberbullying detection under varying supervision levels, (ii) what role parameter-efficient fine-tuning (LoRA) plays in performance improvement, and (iii) whether auxiliary contextual signals, specifically \textit{predicted aggression labels}, can enhance model reasoning and prediction accuracy. To answer these questions, we conduct a systematic evaluation using five publicly available aggression datasets and one cyberbullying dataset. The methodology comprises two main components:\\\\
(1) A comparative baseline evaluation using zero-shot, few-shot, LoRA-based supervised fine-tuning, and joint multi-task learning (MTL), which leverages hard parameter sharing using joint loss weight updates to both adapters.\\\\
(2) An enriched prompt pipeline (EPP) that incorporates predicted aggression labels into cyberbullying classification. 
% The overall process is illustrated in \textbf{Figure~\ref{fig:your_label}}, which shows the system architecture diagram.
%The overall process is illustrated in Fig. 1.

\subsection{Datasets}
Table~\ref{tab:dataset_summary} presents the datasets used for the aggression (D1–D5) and cyberbullying (D6) detection tasks. These identifiers (D1–D6) are used throughout the paper to refer to each dataset. The aggression datasets were collected from social media platforms, including Twitter and online forums, and annotated for varying levels of aggressive behaviour~\cite{rawat2023modelling, nafis2023towards,kumar2018benchmarking, samghabadi2020aggression, kumar2020evaluating}. The cyberbullying dataset (D6) was obtained from a publicly available Kaggle repository and consists of tweets annotated for various forms of online abuse~\cite{fati2025enhancing}.

\begin{figure*}[!t]
    \centering
    \includegraphics[width=\linewidth]{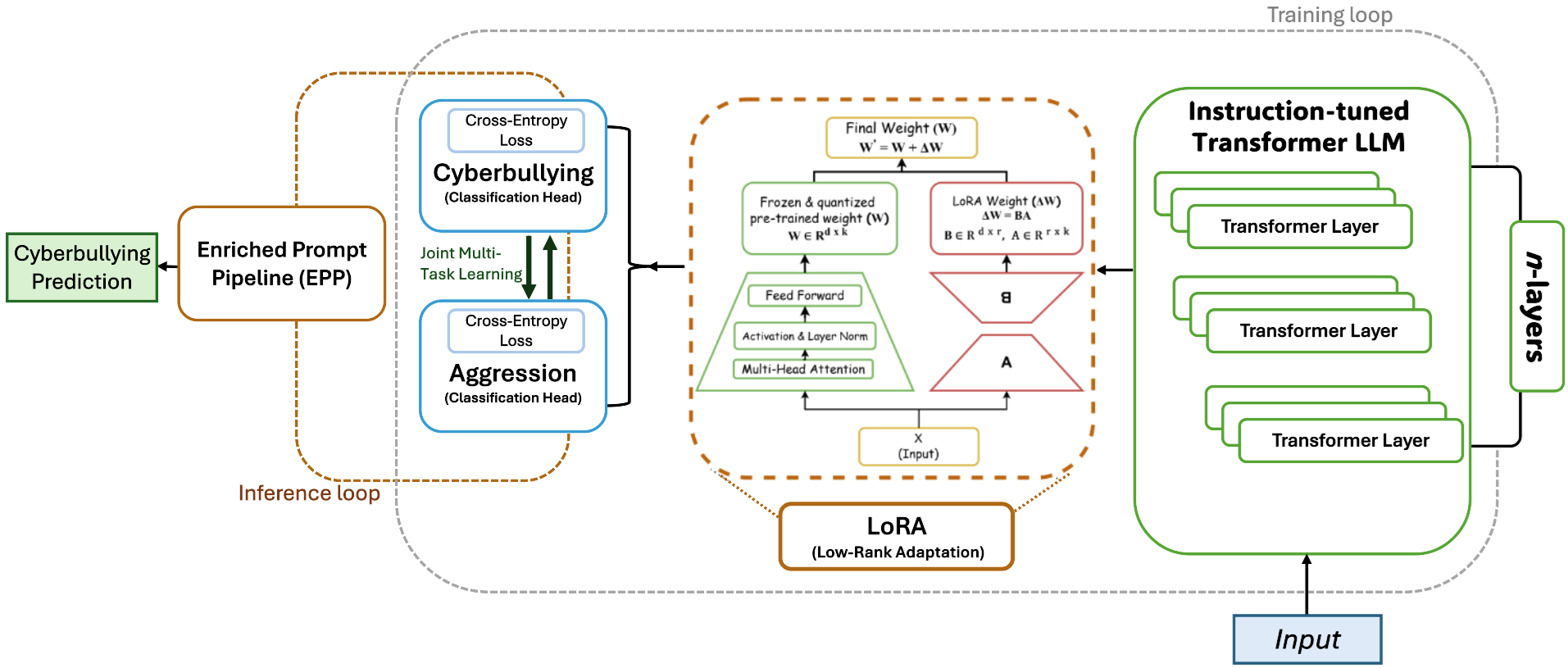}
    \caption{An overview of the proposed system architecture for cyberbullying detection. The diagram illustrates the training loop, which incorporates a LoRA adapter and a joint multi-task learning framework for aggression and cyberbullying, as well as the inference loop that utilises the Enriched Prompt Pipeline (EPP).}
    \label{fig:architecture}
\end{figure*}

\begin{table*}[htbp]
    \centering
    \captionsetup{labelfont=bf, textfont=normalfont}
    \caption{Description of aggression and cyberbullying datasets}
    \label{tab:dataset_summary}
    \small
    \resizebox{0.9\linewidth}{!}{%
    \begin{tabular}{@{} p{3cm} p{6.5cm} p{5.5cm} @{}}
        \toprule
        & \textbf{Aggression Datasets (D1-D5)} & \textbf{Cyberbullying Dataset (D6)} \\
        \midrule

        \textbf{Dataset Name} &
        \begin{tabular}[t]{@{}l@{}}
            D1: Political Aggression Dataset \\
            D2: Code-Mixed Aggression and Offensive \\ \hspace{0.5em} Language Dataset (TRAC - 2) \\
            D3: Aggression Identification\\
            D4: Trac - 2020 Submission \\
            D5: Aggression Identification
        \end{tabular} &
        Cyberbullying Classification \\
        % \addlinespace % Adds a little vertical space between rows
        \hline
        \addlinespace 
        \textbf{Number of instances} &
        \begin{tabular}[t]{@{}l@{}}
            D1: 1,999 samples\\
            D2: 10,000 samples\\
            D3: 12,246 samples\\
            D4: 6,000 samples \\
            D5: 10,000 samples
        \end{tabular} &
        47,000 samples \\
        \hline
        \addlinespace

        \textbf{Categories} &
        \begin{tabular}[t]{@{}l@{}}
            - Not Aggressive (NAG) – Label: 0\\
            - Covertly Aggressive (CAG) – Label: 1\\
            - Overtly Aggressive (OAG) – Label: 2
        \end{tabular} &
        \begin{tabular}[t]{@{}l@{}}
            - Ethnicity/Race\\
            - Religion\\
            - Gender/Sexual\\
            - Not Cyberbullying
        \end{tabular} \\
        \hline
        \addlinespace

        \textbf{Language} &
        \begin{tabular}[t]{@{}l@{}}
            D1: English \\
            D2: Hindi-English (Code-Mixed) \\
            D3: English with Hindi transliteration \\
            D4: English, Hindi, Bengali \\
            D5: English
        \end{tabular} &
        English \\
        \hline
        \addlinespace

        \textbf{Splits} &
        \begin{tabular}[t]{@{}l@{}}
            80\% Training \\
            10\% Test \\
            10\% Validation
        \end{tabular} &
        \begin{tabular}[t]{@{}l@{}}
            75\% Training \\
            25\% Test and validation \\
            (with 2,000 instances in validation)
        \end{tabular} \\
        \bottomrule
    \end{tabular}%
    }
\end{table*}

\subsection{Comparative Evaluation: Baselines}
The first phase of experimentation focuses on establishing baselines using different learning strategies that reflect varying degrees of task supervision and model adaptation complexity. These strategies were applied in a selection of instruction-tuned LLMs, each independently assessed to measure its response to different configurations.\\\\
\textbf{Zero-shot} In this setting, pre-trained instruction-tuned LLMs are evaluated directly on the cyberbullying and aggression detection tasks without task-specific fine-tuning. \\\\
\textbf{Few-shot} This setup introduces a limited set of manually labelled examples to train the model. The objective is to determine the extent to which minimal supervision improves classification, simulating real-world low-resource scenarios and testing the model's rapid learning capacity.\\\\
\textbf{Supervised Fine-Tuning (SFT w/ LoRA)} LoRA is employed to adapt LLMs by injecting trainable low-rank matrices into existing model weights using SFT. Each LoRA is fine-tuned independently on the aggression and cyberbullying tasks. This approach offers a parameter-efficient method for task-specific tuning, ideal for scenarios where computational resources or data volume are constrained.\\\\
\textbf{Joint Multi-Task Learning (MTL)} This strategy fine-tunes the model concurrently on both tasks, aggression and cyberbullying detection, sharing representational layers to potentially capture generalisable features. It tests whether task co-learning facilitates transfer learning benefits or introduces representational conflicts. We leverage multi-task learning by updating model weights using the joint loss (sum) from different heads to both LoRA adapters, and training them with individual classification heads for each task\footnote{\href{https://github.com/center-for-humans-and-machines/transformer-heads}{Transformer Heads Library}}, using embeddings from the final LLM layer.\\\\
These baseline strategies are evaluated using four standard metrics: accuracy, precision, recall, and F1-score. Given the imbalanced, multi-class nature of the cyberbullying detection problem, F1-score is designated as the primary performance measure. This baseline evaluation provides the foundational evidence to motivate more advanced integration strategies.
\subsection{Enhanced Prompt-Pipeline (EPP)}
To further enhance model understanding and classification accuracy, we propose a new methodology that augments model inputs for cyberbullying detection with contextual information derived from aggression classification. This pipeline is based on the hypothesis that aggression cues, \textit{whether overt, covert, or absent, can provide auxiliary semantic signals that help LLMs better detect subtle or implied forms of cyberbullying}. The proposed pipeline is structured into three sequential steps:\\\\
\textbf{Step 1) Aggression Prediction} Each input text is first passed through an aggression detection model fine-tuned using LoRA. The model assigns the post to one of three categories: Overtly Aggressive (OAG), Covertly Aggressive (CAG), or Not-Aggressive (NAG). These predictions serve as summarised emotional cues, offering a linguistic profile of the text's tone or intent.\\\\
\textbf{Step 2) Prompt Enrichment} The predicted aggression label is embedded directly into the classification prompt for cyberbullying detection. This enrichment mechanism is designed to introduce semantic context that primes the model's attention to potentially harmful expressions. A typical enriched prompt follows the format: \\\\
\textit{``This post was predicted as [Overtly Aggressive, Covertly Aggressive, Not-Aggressive]. Based on this, classify the following content for cyberbullying.''}
This strategy transforms the task from isolated content classification into a contextual reasoning problem, where aggression becomes a guiding signal.\\\\
\textbf{Step 3) Cyberbullying Detection} The enriched prompt and the original user-generated post are then fed into a second LoRA-fine-tuned LLM trained for cyberbullying classification (as shown in Figure~\ref{fig:architecture}). By conditioning the model on the aggression label, we aim to improve its ability to detect indirect or context-dependent forms of cyberbullying.

\subsection{Experiment Setup}
To support the methodology described above, all experiments were carried out using three large language models tuned to instructions: \gemmatwob\footnote{\href{https://huggingface.co/google/gemma-2-2b-it}{google/gemma-2-2b-it}}, \gemmanineb\footnote{\href{https://huggingface.co/google/gemma-2-9b-it}{google/gemma-2-9b-it}}, and \gemmafourb\footnote{\href{https://huggingface.co/unsloth/gemma-3-4b-it-unsloth-bnb-4bit}{unsloth/gemma-3-4b-it-unsloth-bnb-4bit}}. These models were selected primarily to represent a range of parameter sizes and fine-tuning capabilities. For standard LoRA fine-tuning, each model was trained for 1 epoch, reflecting a lightweight and computationally efficient adaptation process. In the joint multi-task learning configuration, training duration was extended to between 3 and 6 epochs, allowing the models to jointly optimise across tasks. Training was performed with a batch size of 8 and a learning rate set to \texttt{1e-4}. The LoRA rank parameter was fixed at $r = 8$, enabling low-rank updates without full-scale retraining. Few-shot evaluations were conducted using $k = 3$ examples per class, simulating limited supervision scenarios. Execution times varied depending on model size and architecture. Smaller models completed training in approximately 45 minutes, while larger models required up to 9 hours. All experiments were executed using two GPUs equipped with 24GB VRAM.
Model evaluation employed four performance metrics: accuracy, precision, recall, and F1-score. Given the inherent class imbalance in cyberbullying datasets, the macro-averaged F1-score was treated as the primary performance indicator throughout all evaluation phases~\cite{luo2025reinforced}.

\begin{table*}[tbp]
    \centering
    \captionsetup{labelfont=bf, textfont=normalfont}
    \caption{\textit{macro}-F1 score comparison across models and methods for aggression and cyberbullying detection. The table groups results from zero-shot, few-shot, LoRA, Multi-Task Learning (MTL), and Enriched Prompt Pipeline (EPP) evaluations. \textbf{Bold} indicates the best performing method within a specific comparison for a given task, while the \textit{italicized} value is noted for comparison against LoRA.}
    \label{tab:all_comparisons_unified}
    
    \resizebox{1\linewidth}{!}{%
    \begin{tabular}{@{} l ccccc ccccc @{}}
        \toprule
        \multirow{2}{*}{\textbf{Model}} & \multicolumn{5}{c}{\textbf{Aggression Detection}} & \multicolumn{5}{c}{\textbf{Cyberbullying Detection}} \\
        \cmidrule(lr){2-6} \cmidrule(lr){7-11}
        & Zero-shot & Few-shot & LoRA & MTL & EPP & Zero-shot & Few-shot & LoRA & MTL & EPP \\
        \midrule
        
        % Gemma-2-2B Data
        \gemmatwob & 0.54 & 0.56 & \textbf{0.67} & 0.51 & \textbf{0.67} & 0.63 & \textbf{0.83} & 0.84 & 0.90 & \textbf{0.99} \\
        \addlinespace

        % Gemma-2-9B Data
        \gemmanineb & 0.57 & 0.60 & 0.65 & 0.53 & 0.65 & 0.79 & \textbf{0.83} & \textbf{0.93} & \textit{0.89} & \textbf{0.99} \\
        \addlinespace

        % Gemma-3-4B Data
        \gemmafourb & 0.53 & \textbf{0.63} & 0.50 & 0.49 & 0.50 & 0.34 & 0.57 & 0.84 & 0.76 & 0.86 \\

        \bottomrule
    \end{tabular}%
    }
\end{table*}
\section{Results And Discussion}\label{sec:Results}
The selected approaches encompass a range of generalisation, supervision, and efficiency scenarios to provide a comprehensive evaluation: 
Zero-shot learning evaluates a model’s baseline capacity to perform a task with no task-specific supervision. Few-shot learning tests the model's adaptability to small, annotated datasets. LoRA enables efficient adaptation by injecting trainable low-rank updates. MTL leverages shared representation learning across related tasks. The Enriched Prompt Pipeline introduces auxiliary contextual knowledge to improve downstream task comprehension.

\subsection{Zero-shot vs. Few-shot Performance}

In the zero-shot setting, models are tested without task-specific examples to assess generalisation from instruction tuning alone. Few-shot learning introduces limited supervision to evaluate the minimum data required for meaningful performance gains. As shown in Table~\ref{tab:all_comparisons_unified}, few-shot learning consistently yields higher F1 scores than zero-shot learning across all three models. For aggression detection, the gemma-3-4b-it-unsloth-bnb-4bit model increases from an F1-score of 0.53 (zero-shot) to 0.63 (few-shot), while the \gemmatwob and \gemmanineb models show smaller increases from 0.54 to 0.56 and 0.57 to 0.60, respectively. These results suggest that models are better able to identify explicit hostility or implicit aggression patterns when supervised in a limited but focused manner. Cyberbullying detection also shows a substantial increase in F1 scores. As detailed in the table, the \gemmatwob model's F1-score increases from 0.63 to 0.83, and \gemmanineb shows a similar increase from 0.79 to 0.83. This indicates that even a few examples help models capture context-dependent and often nuanced instances of cyberbullying, such as sarcasm, coded language, or indirect threats. The results presented in Table~\ref{tab:all_comparisons_unified} thus emphasise the critical role of task-specific samples in enabling LLMs to understand and differentiate harmful content effectively.

\subsection{Few-shot vs. LoRA SFT}

LoRA enables efficient task-specific fine-tuning by updating fewer parameters, making it a viable approach for resource-constrained settings. As detailed in Table~\ref{tab:all_comparisons_unified}, this approach results in a higher F1 score for aggression detection compared to few-shot learning for two of the models: for example, \gemmatwob increases from 0.56 to 0.67 and \gemmanineb from 0.60 to 0.65. However, gemma-3-4b-it-unsloth-bnb-4bit drops from 0.63 (few-shot) to 0.50 (LoRA), likely due to overfitting or misalignment in instruction-heavy or compact models.

In the cyberbullying task, LoRA fine-tuning also results in higher F1 scores for all models. \gemmanineb, for instance, shows a score increase from 0.83 (few-shot) to 0.93 (LoRA), while gemma-3-4b-it-unsloth-bnb-4bit score increases from 0.57 to 0.84. These results indicate that LoRA is beneficial for tasks involving rich semantic complexity, as it enables the models to refine their internal representations without extensive retraining. The results in Table~\ref{tab:all_comparisons_unified} thus highlight the effectiveness of LoRA for adapting models to challenging, multi-dimensional language classification problems, especially in cases where few-shot learning reaches its performance ceiling.

\subsection{LoRA SFT vs. MTL}

Multi-task learning (MTL) investigates whether jointly training on aggression and cyberbullying detection tasks enables models to learn more generalizable linguistic features, based on shared traits such as hostile intent and emotional intensity. As shown in Table~\ref{tab:all_comparisons_unified}, a comparison with LoRA reveals that MTL consistently yields a lower F1 score for aggression detection across all models. For example, \gemmatwob shows a decrease from 0.67 (LoRA) to 0.51 (MTL), and \gemmanineb from 0.65 to 0.53. This drop in performance suggests that joint optimization may cause task interference, especially when there is a disparity in task difficulty or dataset size, potentially leading to a trade-off where one task improves at the expense of the other.

The results of cyberbullying detection under MTL are mixed. \gemmatwob shows an increase in its F1 score from 0.84 to 0.90, suggesting some benefit from joint learning. However, other models like gemma-3-4b-it-unsloth-bnb-4bit and \gemmanineb show a decrease in performance even with more epochs. For instance,  \gemmanineb score decreases from 0.93 (LoRA) to 0.89 (MTL), as highlighted by the italicized value. These outcomes illustrate that MTL's effectiveness is model- and task-dependent, requiring careful tuning to avoid negative transfer.

\subsection{LoRA SFT vs. Proposed EPP}

We introduce the Enriched Prompt Pipeline (EPP) to address limitations of LoRA and MTL by enhancing cyberbullying detection through context engineering (aggression-informed prompts) by sequentially prompting adapters. This approach augments model input with predicted aggression labels, improving classification accuracy without increasing model complexity or training time. As shown in Table~\ref{tab:all_comparisons_unified}, EPP shows a substantial increase in F1 scores for cyberbullying detection: for instance, \gemmatwob increases from 0.84 (LoRA) to 0.99, and \gemmanineb also shows an increase from 0.93 to 0.99. Even the previously less stable gemma-3-4b-it-unsloth-bnb-4bit model shows an increase from 0.84 to 0.86. 

As expected, the aggression results in Table~\ref{tab:all_comparisons_unified} are identical for both the LoRA and EPP methods. This is because no retraining was performed on aggression tasks in the EPP setup. Instead, aggression labels were solely used to enrich the cyberbullying prompts, aligning with the study’s main objective to enhance cyberbullying detection using auxiliary aggression signals.

\section{Conclusion and Future Work}\label{sec:conc}

This study focused on improving cyberbullying detection using LLMs, with aggression detection leveraged as an auxiliary signal to enhance performance. We evaluated multiple learning paradigms, including zero-shot, few-shot, LoRA fine-tuning, joint MTL, and enriched prompting with three instruction-tuned LLMs. Our proposed enriched prompt pipeline, which incorporates aggression-related cues into model inputs, showed substantial gains, demonstrating its utility in improving model generalization. This work indicates that lightweight, context-sensitive prompt augmentation is a promising approach for tasks that require socially sensitive NLP.

Our future work will focus on assessing a wider range of instruction-tuned LLMs, including models such as Llama and Mistral. Although we encountered technical challenges related to accessing and deploying these models during this study, we plan to address these in future experiments.

\clearpage
\bibliographystyle{acl_natbib}
\bibliography{ranlp2025}

\end{document}